\pdfoutput=1

\documentclass[11pt]{article}

\usepackage[final]{acl}

\usepackage{times}
\usepackage{latexsym}

\usepackage[T1]{fontenc}

\usepackage[utf8]{inputenc}

\usepackage{microtype}

\usepackage{inconsolata}
\usepackage{multirow}
\usepackage{microtype}
\usepackage{graphicx}
\usepackage{tabularx}
\usepackage{subfigure}
\usepackage{booktabs} 
\usepackage{amsmath}
\usepackage{xspace}

\usepackage{times}
\usepackage{latexsym}
\usepackage[T1]{fontenc}
\usepackage[utf8]{inputenc}
\usepackage{microtype}
\usepackage{inconsolata}
\usepackage{multirow}
\usepackage{microtype}
\usepackage{graphicx}
\usepackage{tabularx}
\usepackage{subfigure}
\usepackage{booktabs} 
\usepackage{amsmath}
\usepackage{xspace}
\usepackage{algorithm}
\usepackage[noend]{algorithmic}
\usepackage{colortbl} 
\usepackage{xcolor}
\usepackage{wrapfig}
\usepackage{subcaption}

\usepackage[utf8]{inputenc} 
\usepackage[T1]{fontenc}    
\usepackage{hyperref}       
\usepackage{url}            
\usepackage{booktabs}       
\usepackage{amsfonts}       
\usepackage{nicefrac}       
\usepackage{microtype}      
\newcommand{\ours}{SIMA\xspace}

\usepackage{soul}

\usepackage[export]{adjustbox}

\usepackage{amsmath,amsfonts,bm}

\def\eqref#1{equation~\ref{#1}}

\def\1{\bm{1}}

\DeclareMathAlphabet{\mathsfit}{\encodingdefault}{\sfdefault}{m}{sl}
\SetMathAlphabet{\mathsfit}{bold}{\encodingdefault}{\sfdefault}{bx}{n}

\def\gD{{\mathcal{D}}}

\def\gL{{\mathcal{L}}}

\newcommand{\E}{\mathbb{E}}

\title{Enhancing Visual-Language Modality Alignment in Large Vision Language Models via Self-Improvement}

\author{%
  Xiyao Wang$^{1, 3 \dag}$\thanks{The work is partially done during Xiyao Wang’s internship at GE Healthcare. }, Jiuhai Chen$^{1}$, Zhaoyang Wang$^{2}$, Yuhang Zhou$^{1}$, Yiyang Zhou$^{2}$ \\
  \textbf{Huaxiu Yao$^{2}$}, \textbf{Tianyi Zhou$^{1}$}, \textbf{Tom Goldstein$^{1}$}, \textbf{Parminder Bhatia$^{3}$},  \textbf{Taha Kass-Hout$^{3}$} \\
  \textbf{Furong Huang$^{1 \ddag}$}, \textbf{Cao Xiao$^{3 \ddag}$} \\
  $^1$University of Maryland \quad $^2$UNC-Chapel Hill \quad $^3$GE Healthcare \\
  $^\dag$\texttt{xywang@umd.edu} $^\ddag$ Equal advising\\
}
  
\begin{document}
\maketitle
\begin{abstract}

Large vision-language models (LVLMs) have achieved impressive results in visual question-answering and reasoning tasks through vision instruction tuning on specific datasets. However, there remains significant room for improvement in aligning visual and language modalities. Existing methods often depend on external models or data, leading to uncontrollable and unstable alignment results. In this paper, we propose SIMA, a self-improvement framework that enhances visual and language modality alignment without external dependencies. SIMA leverages existing vision instruction tuning datasets to self-generate responses, incorporating an in-context self-critic mechanism that constructs preference pairs for tuning. Crucially, our approach allows LVLMs to act as critics by designing effective critic prompts, eliminating the need for additional fine-tuning with external instruction data. We introduce three novel visual metrics within the self-critic process to guide judgement, significantly improving the accuracy of self-critic. Through extensive experiments across 14 hallucination and comprehensive benchmarks, we demonstrate that SIMA significantly improves LVLM's performance and outperforms previous approaches, achieving superior modality alignment.

\end{abstract}

\begin{figure}[!ht]
\centering
\includegraphics[width=0.5\textwidth]{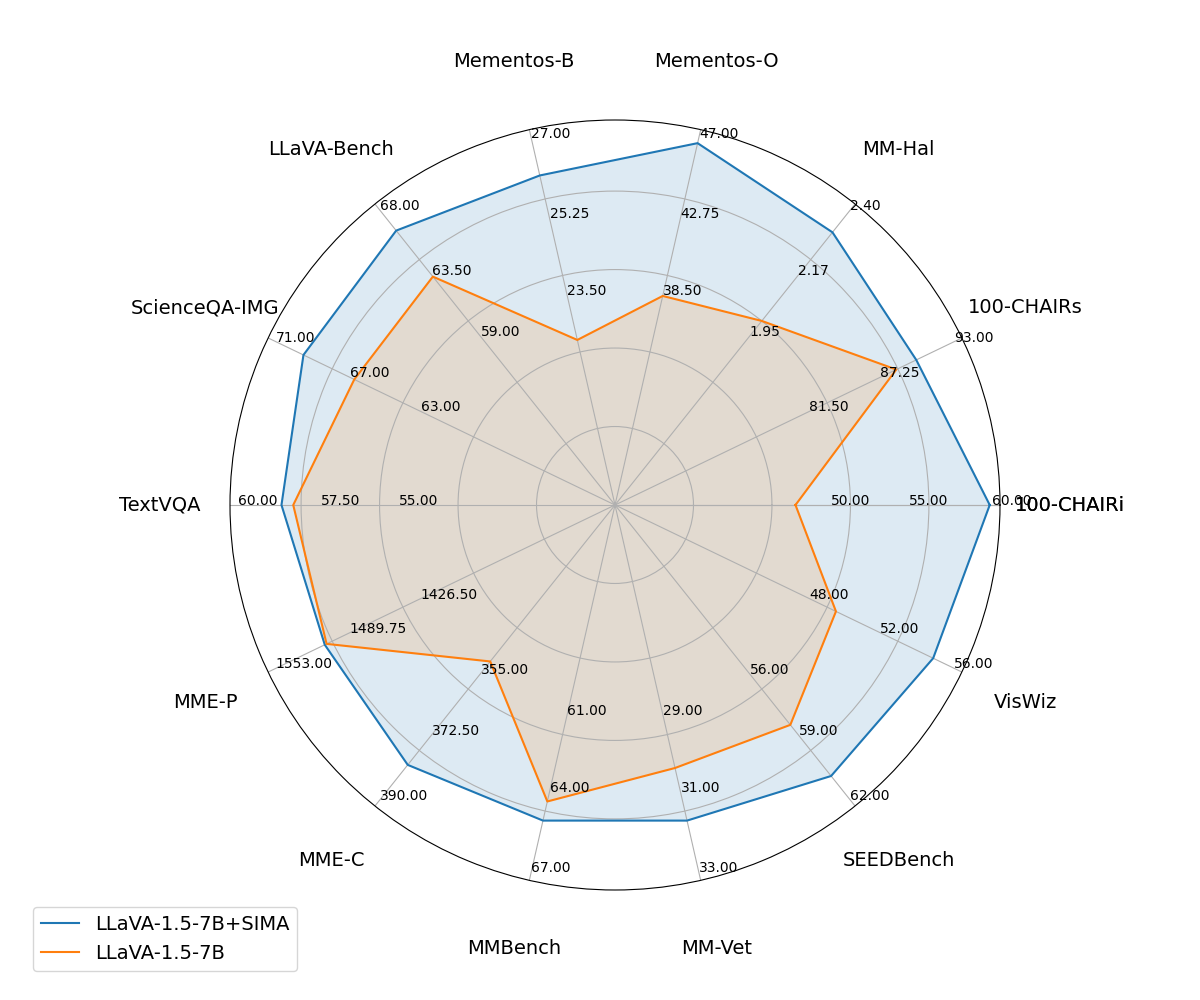}
\caption{Performance comparison between our propose framework \ours and LLaVA-1.5-7B on 14 hallucination and comprehensive benchmarks. After applying \ours, LLaVA's performance is improved significantly across all benchmarks, with an average performance increase of 7.5\%.}
\label{fig:main_exp}
\end{figure}

\begin{figure}[!ht]
\centering
\includegraphics[width=0.4\textwidth]{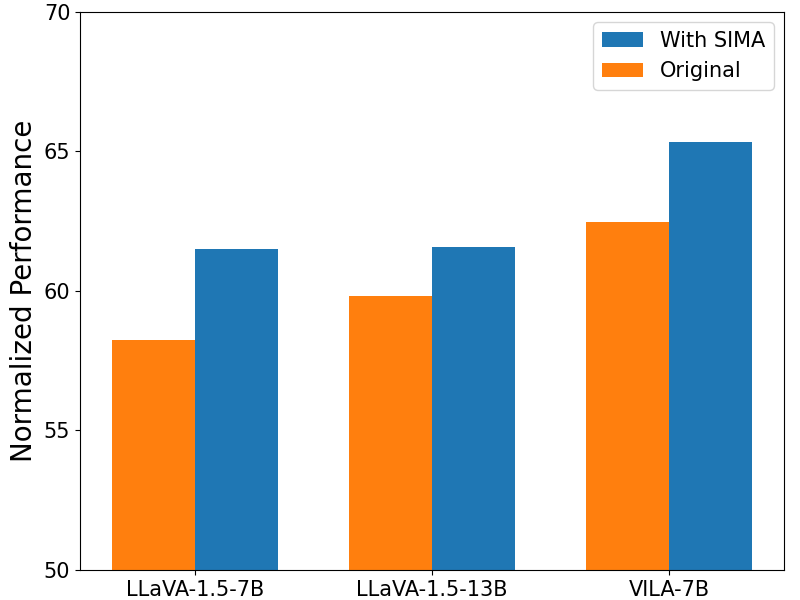}
\caption{Normalized average performance across 14 hallucination and comprehensive benchmarks of three different LVLMs before and after using \ours. \ours demonstrates significant improvement on all three LVLMs.}
\label{fig:norm_all_main}
\end{figure}

\section{Introduction}

\begin{figure*}[!ht]
\centering
\includegraphics[width=1.01\textwidth]{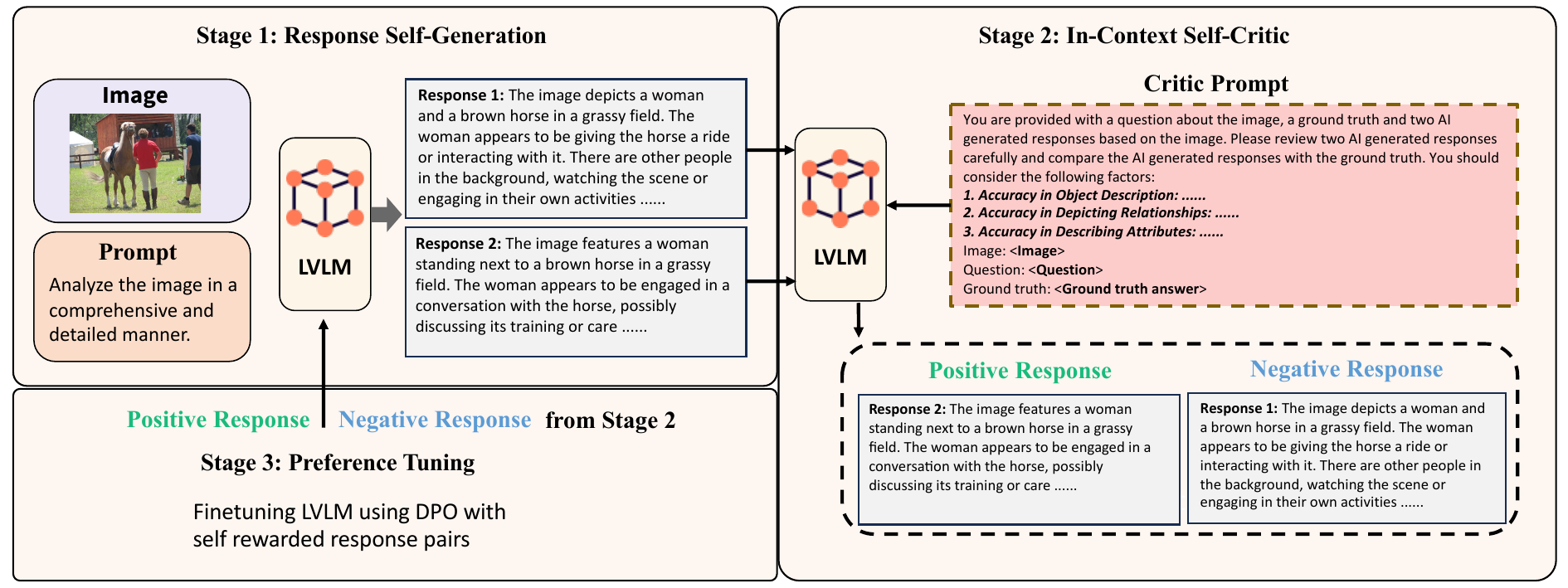}
\caption{Flowchart of the \ours framework. \ours consists of three parts: Response Self-Generation, In-Context Self-Critic, and Preference Tuning.}
\label{fig:main}
\end{figure*}

Large Language Models (LLMs) \cite{brown2020language, chowdhery2023palm, touvron2023llama} have significantly advanced the development of Large Vision Language Models (LVLMs) through pre-training on image-text pairs \cite{alayrac2022flamingo, xiao2023florence} or fine-tuning on specialized vision instruction datasets \cite{liu2023improved, liu2024visual, zhu2023minigpt}. Despite these advancements, effectively aligning visual and language modalities remains a critical challenge in LVLMs.

Recent works~\cite{sun2023aligning, zhao2023beyond, zhou2024aligning} have attempted to enhance this alignment through preference tuning methods such as reinforcement learning from human feedback (RLHF) \cite{ouyang2022training} and Direct Preference Optimization (DPO) \cite{rafailov2024direct}. However, these methods often rely on external models or human-labeled data, introducing issues of uncontrollable and unstable alignment results. Specifically, they face two major challenges: \textbf{(1) Distribution Shifts:} Utilizing external LVLMs to generate preference pairs can introduce hallucinations from external models that are not representative of the current model's inference behavior \cite{li2023silkiepreferencedistillationlarge, zhao2023beyond, zhou2024aligning}. This discrepancy can lead to instability in the optimization process and potentially degrade performance. \textbf{(2) High Costs:} Dependence on human-labeled datasets or feedback from third-party AI models incurs significant annotation or API costs, making it difficult to scale high-quality preference datasets in resource-constrained environments \cite{sun2023aligning, yu2024rlhf, yu2024rlaif, xiyao2024scaling}.


To address these challenges, we propose the \textbf{Self-Improvement Modality Alignment (\ours)} framework, designed to enhance the alignment between visual and language modalities within LVLMs through a self-improvement mechanism. \ours eliminates the need for external data or models by leveraging the intrinsic capabilities of the model itself to generate diverse responses. Moreover, it utilizes the model's own judgment for evaluating response quality, thus avoiding the high costs associated with external feedback and scaling up preference datasets efficiently.


\ours consists of three stages: \textbf{response self-generation}, \textbf{in-context self-critic}, and \textbf{preference tuning}. In the response self-generation stage, we sample prompts from the current LVLM's visual instruction tuning dataset to generate diverse responses without introducing external data or models. During the in-context self-critic stage, a carefully designed critic prompt allows the LVLM to evaluate all self-generated responses and form preference pairs. Finally, preference tuning is applied to update the LVLM based on these pairs.


The core innovation of SIMA lies in the in-context self-critic process, which offers several key advantages: \textbf{(1) Self-Critic without Fine-Tuning:} Unlike previous self-rewarding methods in LLMs that require additional instruction tuning before the critic step \cite{yuan2024self, pang2024iterativereasoningpreferenceoptimization, wu2024metarewardinglanguagemodelsselfimproving}, our approach shows that by properly configuring the critic prompt, the LVLM can accurately evaluate responses without fine-tuning. \textbf{(2) Visual Critic Metrics:} To ensure accurate evaluation of self-generated responses, we introduce three visual critic metrics within the prompt—Accuracy in Object Description, Accuracy in Depicting Relationships, and Accuracy in Describing Attributes—each contributing to a more precise evaluation of visual content.


We apply \ours to LLaVA-1.5~\cite{liu2023improved} and VILA~\cite{lin2024vila}, evaluating it across 14 hallucination and comprehensive benchmarks. The experimental results show that \ours not only mitigates hallucinations but also significantly enhances comprehension capabilities in LVLMs. As illustrated in Figure~\ref{fig:main_exp}, the performance of LLaVA-1.5-7B, LLaVA-1.5-13B, and VILA-7B improved by 7.5\%, 4.5\%, and 5.3\%, respectively.
Additionally, our method outperforms other preference-tuning approaches that rely on external models and data.

\textbf{The contribution of this paper can be summarized as follows:}
\textbf{(1)} We introduce Self-Improvement Modality Alignment \ours, a novel framework designed to enhance alignment between visual and language modalities in LVLMs. To the best of our knowledge, \ours is the first to achieve self-improvement in LVLMs without external data or third-party AI models.
\textbf{(2)}  We propose the in-context self-critic method, enabling LVLMs to accurately evaluate responses without instruction tuning, significantly improving judgment accuracy through three visual critic metrics.
\textbf{(3)} \ours demonstrates significant performance improvements in LLaVA-1.5-7B, LLaVA-1.5-13B and VILA-7B on 14 hallucination and comprehensive benchmarks, validating the effectiveness of our approach.


\section{Self-Improvement Modality Alignment}

In this section, we introduce the proposed \textbf{S}elf-\textbf{I}mprovement \textbf{M}odality \textbf{A}lignment (\ours) framework. 
\ours is consisted of three stages: response self-generation, in-context self-critic, and preference tuning. 
We will first explain how to obtain self-generated response candidates in Sec~\ref{sec: stg1}, then discuss how to use model itself $\pi_\theta$ to critique the response candidates in Sec~\ref{sec: stg2}. Finally, we will introduce how to use self-rewarded responses to update the $\pi_\theta$ in Sec~\ref{sec: stg3}.
The pseudo-code of \ours is provided in Algorithm~\ref{alg: SIMA}.

\begin{algorithm}{}
\caption{\ours}
\label{alg: SIMA}
\begin{algorithmic}[1]
\REQUIRE{
Prompt Dataset $\{ x_i, I_i \}_{i \in [N]}$, Preference dataset $\gD_p=\{ \}$ , Current optimized LVLM \textcolor{blue}{$\pi_\theta$}, Reference model \textcolor{blue}{{}$\pi_{ref}$}}, 
\FOR{$i = 1, \ldots, N$}
    \STATE \textcolor[RGB]{230,145,145}{\textit{// Stage 1: Response self-generation}}
    \STATE Generate one response using greedy decoding with \textcolor{blue}{$\pi_\theta$},
    \STATE Generate one response using temperature sampling with \textcolor{blue}{$\pi_\theta$},
    \STATE \textcolor[RGB]{230,145,145}{\textit{// Stage 2: In-context self-critic}}
    \STATE Criticizing two generated responses with \textcolor{blue}{$\pi_\theta$},
    \STATE Add preference pair $\{y_w, y_l\}$ into $\gD_p$,
\ENDFOR
\STATE \textcolor[RGB]{230,145,145}{\textit{// Stage 3: Preference tuning}}
\STATE Update \textcolor{blue}{$\pi_\theta$} using Eq~\ref{eq: dpo} with \textcolor{blue}{{}$\pi_{ref}$}
\end{algorithmic}
\end{algorithm}


\begin{figure*}[!ht]
\centering
\includegraphics[width=\textwidth]{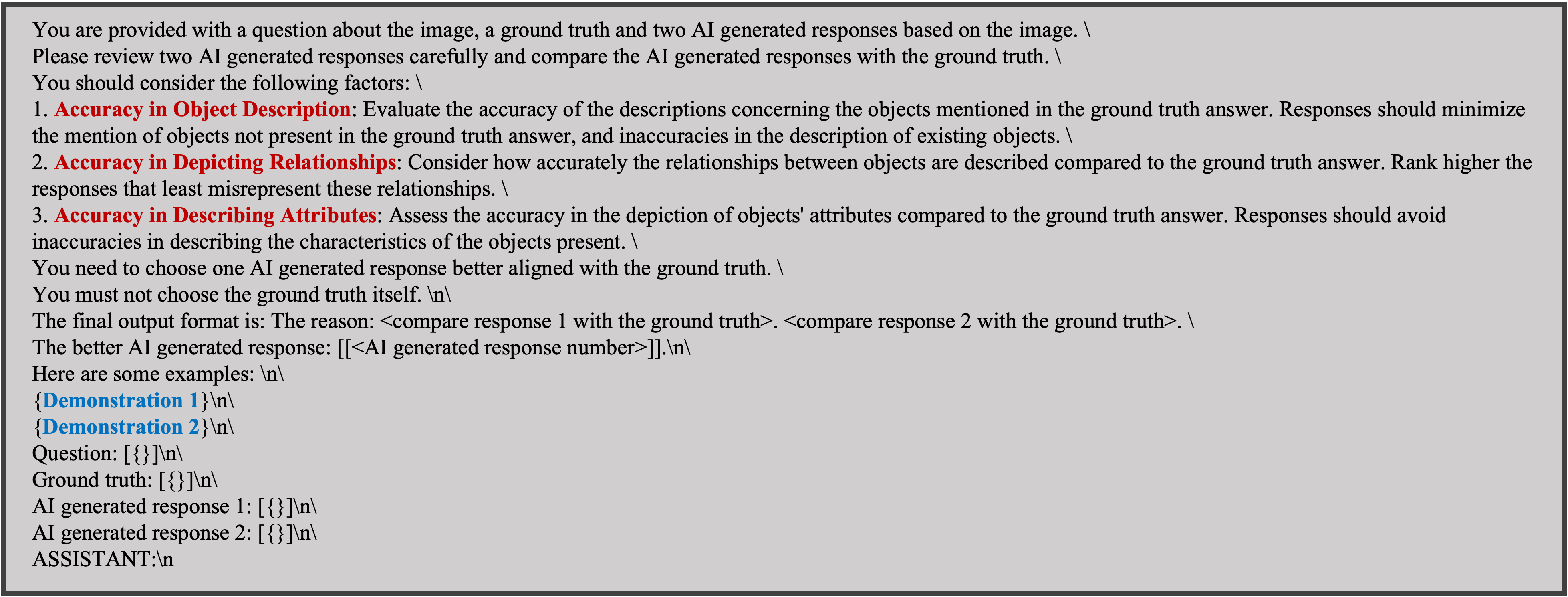}
\caption{Critic prompt structure used for in-context self-critic.}
\label{fig: critic prompt demo}
\vspace{-10pt}
\end{figure*}

\subsection{Response self-generation}
\label{sec: stg1}

Previous works often require the introduction of external models to generate preference dataset to improve current LVLM \citep{sun2023aligning, zhou2024aligning}. 
However, due to the significant distribution shift between the external models and the currently optimized LVLM, 
the generated dataset by these approaches may not be helpful to the LVLM.
For example, a common method to obtain negative responses is to use external models to deliberately modify the ground truth and inject object hallucinations \citep{zhou2024aligning}, while the hallucinations generated by external models do not necessarily indicate that the currently optimized model would produce. 
In this case, using such data for learning can not enhance LVLM. 

Based on our goal to identify and correct the potential misunderstandings the current LVLM may have about images and improve the modality alignment, we propose using the currently optimized LVLM to self-generate responses. 
This approach avoids the potential distribution shift introduced by external models. 
As shown in Stage 1 of Figure~\ref{fig:main}, given an image and its corresponding prompt, we use the currently optimized model to generate two different response candidates for subsequent ranking and preference tuning. Specifically, the two responses are generated using \textbf{greedy decoding} and \textbf{temperature sampling} to ensure diversity between the responses.


\subsection{In-context self-critic}
\label{sec: stg2}

The core part of \ours is criticizing the self-generated responses without introducing an additional reward model. As shown in Stage 2 of Figure~\ref{fig:main}, we directly input the self-generated responses and the critic prompt into the currently optimized LVLM. The LVLM then selects the better response as the positive response and the other one as the negative response. The most critical part of this stage is designing an appropriate critic prompt, since the quality of the critic directly determines the performance of the LVLM optimized using the response pairs. If the worse response is selected as the positive response, it will harm the training of the LVLM.

Our critic prompt consists of the following parts:
\begin{itemize}
    \item \textbf{Image, Question, and Ground Truth Response:} Unlike LLMs, which primarily focus on aspects such as the format, helpfulness, and harmlessness of the textual response, LVLMs primarily focus on the accuracy of the response's understanding of the image content. This means there is a quantifiable accuracy metric to measure the quality of the response. Therefore, during in-context self-critic, we must provide the ground truth response as a reference to choose the positive response. It is worth noting that since the prompts used to generate responses are sampled from the training data of the visual instruction tuning stage, the corresponding ground truth responses have all been used for visual instruction tuning. Hence, using the ground truth in the in-context self-critic stage is reasonable.

    \item \textbf{Three critic metrics:} Although we provide the ground truth response as a reference, without proper guidance, the LVLM might still choose a response that aligns more with the ground truth in terms of output format or harmlessness rather than focusing on the accuracy of visual comprehension. Therefore, we propose three metrics to guide LVLM ranking, ensuring it select the positive response from the visual comprehension perspective. The three critic metrics are: \textbf{Accuracy in Object Description}, \textbf{Accuracy in Depicting Relationships}, and \textbf{Accuracy in Describing Attributes}.

    Accuracy in Object Description aims to guide current LVLM in evaluating the accuracy of the descriptions concerning the objects mentioned in the ground truth answer. The responses should minimize the mention of objects not present in the ground truth answer and inaccuracies in the description of existing objects.
    Accuracy in Depicting Relationships considers how accurately the relationships between objects are described compared to the ground truth answer and aims to let LVLM rank higher the responses that least misrepresent these relationships.
    Accuracy in Describing Attributes assesses the accuracy in depicting objects' attributes compared to the ground truth answer. The responses should avoid inaccuracies in describing the characteristics of the objects present.
    
    \item \textbf{Demonstrations:} To ensure the correct format of the ranking output, we also leverage in-context learning by providing two ranking demonstrations in the designed ranking prompt for the LVLM to imitate.
\end{itemize}

In Figure~\ref{fig: critic prompt demo}, we provide the structure of the critic prompt. For the detailed critic prompt, please refer to the Appendix~\ref{App1: Detailed ranking prompt}.

\subsection{Preference tuning}
\label{sec: stg3}

After obtaining the preference pairs through self-ranking, we use these preference pairs to perform preference tuning on the current LVLM. We choose direct preference optimization (DPO)~\citep{rafailov2024direct} as the preference tuning method.
The preference dataset is denoted as $\gD_p=\{(I, x, y_w, y_l)\}$, where $I$ is the image, $x$ is the corresponding question, $y_w$ is the positive response and $y_l$ is the negative response,
the DPO objective is defined as below:
\begin{equation}\label{eq: dpo}
\resizebox{\hsize}{!}{$
\begin{aligned}
\gL_{DPO}(\pi_\theta; \pi_{\text{ref}}) = -\E_{(x, y_u, y_l) \sim \gD} [ \log \sigma ( & \beta \log \frac{\pi_\theta(y_w | x, I)}{\pi_{\text{ref}}(y_w | x, I)} \\
& - \beta \log \frac{\pi_\theta(y_l | x, I)}{\pi_{\text{ref}}(y_l | x, I)} ) ],
\end{aligned}
$
}
\end{equation}
where $\pi_\theta$ is the current optimized LVLM and $\pi_{\text{ref}}$ is the base reference model, both models are initialized with visual instruction tuning weights.
$\sigma$ is the logistic function.

\section{Experiment}

In this section, we conduct experiments and aim to answer the following questions: 1. How much does \ours improve baseline performance? 2. How significant are the three ranking metrics in the ranking prompt?
 
\subsection{Benchmark evaluation}
\label{sec: benchmark eval}

\begin{table*}[!htb]
\setlength{\abovecaptionskip}{0.2cm}
\centering
\caption{Performance comparison between \ours and other baselines on hallucination benchmarks}
\label{exp: hallucination}
\resizebox{0.77\textwidth}{!}{ 
\begin{tabular}{l|ccccc}
\toprule 
LVLMs & CHAIRs $\downarrow$ & CHAIRi $\downarrow$ & MM-Hal $\uparrow$ & Mementos$^{\mathrm{O}}$ $\uparrow$ & Mementos$^{\mathrm{B}}$ $\uparrow$ \\
\midrule
LLaVA-1.5-7B              & 50.8     & 11.7    & 2.04 & 39.29\% & 23.02\% \\
+ RLHF                    & 45.3     & 11.1    & 2.11 & 40.53\% & 22.71\% \\
+ GT-DPO                  & 47.3     & 11.2    & 2.00 & 43.67\% & 24.35\% \\
+ HA-DPO                  & 46.5     & 10.7    & 1.97 & 41.07\% & 23.58\% \\
+ POVID                   & 48.4     & 11.3    & 2.28 & 42.95\% & 23.84\% \\
\rowcolor{gray!30}
+ \textbf{\ours} (ours)   & \textbf{40.9}     & \textbf{10.4}    & \textbf{2.30} & \textbf{46.08\%} & \textbf{26.03\%} \\
\midrule
LLaVA-1.5-13B             & 48.6     & 10.8   & 2.19    & 40.37\% & 24.65\% \\
+ GT-DPO                  & 47.2     & 10.8   & 2.27    & 42.59\% & 25.84\% \\
\rowcolor{gray!30}
+ \textbf{\ours} (ours)   & \textbf{45.8}     & \textbf{10.6}   & \textbf{2.41} & \textbf{45.84\%} & \textbf{27.17\%} \\
\midrule
VILA-7B                   & 34.7	    &  9.2	   &2.53	   & 41.96\%	          &25.88\% \\
+ GT-DPO                  & 32.4        &  8.9     &2.61       & 44.25\%          &26.91\% \\
\rowcolor{gray!30}
+ \textbf{\ours} (ours)   & \textbf{28.4}	&\textbf{8.4}	&\textbf{2.66}	&\textbf{48.15\%}	&\textbf{27.04\%} \\
\bottomrule
\end{tabular}
}
\end{table*}

\begin{table*}[!htb]
\setlength{\abovecaptionskip}{0.2cm}
\centering
\caption{Performance comparison between \ours and other baselines on comprehensive benchmarks}
\label{exp: comprehensive}
\resizebox{\textwidth}{!}{ 
\begin{tabular}{l|ccccccccc}
\toprule 

LVLMs & LLaVA$^{\mathrm{W}}$ $\uparrow$ & SQA$^{\mathrm{I}}$ $\uparrow$ & VQA$^{\mathrm{T}}$ $\uparrow$ & MME$^{\mathrm{P}}$ $\uparrow$ & MME$^{\mathrm{C}}$ $\uparrow$ & MMB $\uparrow$ & MM-Vet $\uparrow$ & SEED $\uparrow$ & VisWiz $\uparrow$ \\
\midrule
LLaVA-1.5-7B              & 63.4 & 66.8 & 58.2 & 1506.4 & 355.7 & 64.3 & 30.5 & 58.6 & 50.0\\
+ RLHF                    & 63.7 & 65.8 & 58.3 & 1508.2 & 360.2 & 60.4 & 31.1 & 60.0 & 52.2  \\
+ GT-DPO                  & 64.7 & 67.4 & 58.1 & \textbf{1510.8} & 365.0 & 64.6 & 31.2 & 60.4 & 53.8\\
+ HA-DPO                  & 64.2 & 68.1 & 58.0 & 1507.2 & 362.3 & 63.9 & 30.9 & 60.2 & 53.9\\
+ POVID                   & 65.3 & \textbf{69.2} & 58.1 & 1493.5 & 363.5 & 64.1 & 31.3 & 60.3 & 54.0\\
\rowcolor{gray!30}
+ \textbf{\ours} (ours)   & \textbf{66.1} & 69.1 & \textbf{58.5} & 1507.7 & \textbf{379.3} & \textbf{64.9} & \textbf{31.6} & \textbf{60.6} & \textbf{54.4}\\
\midrule
LLaVA-1.5-13B             & 66.5  & 71.6 & \textbf{61.3} & 1531.1 & 296.1 & 67.7 & 36.1 & 61.6 & 53.6\\
+ GT-DPO                  & 66.9  & 72.3 & 61.2          & 1532.6 & 296.7 & 68.0 & 36.3 & 62.2 & 54.4\\
\rowcolor{gray!30}
+ \textbf{\ours} (ours)   & \textbf{67.4}  & \textbf{72.5} & 61.2 & \textbf{1538.1} & \textbf{298.6} & \textbf{68.4} & \textbf{38.3} & \textbf{63.0} & \textbf{55.5} \\
\midrule
VILA-7B                   & 69.7	&68.2	&64.4	&1533.0	&316.4	&68.9	&34.9	&61.1	&57.8\\
+ GT-DPO                  & 71.4    & 70.6  & 65.9  & 1547.8    & 325.7 &69.0   &37.1   &61.9   &60.3 \\
\rowcolor{gray!30}
+ \textbf{\ours} (ours)   & \textbf{73.5}	&\textbf{72.2}	&\textbf{66.1}	&\textbf{1559.6}	&\textbf{326.8}	&\textbf{69.2}	&\textbf{38.4}	&\textbf{62.5}	&\textbf{62.1} \\
\bottomrule
\end{tabular}
}
\end{table*}

\paragraph{Implementation details} 
Since LLaVA~\citep{liu2024visual} is the most widely used open-source LVLM and following recent LVLM preference tuning studies \citep{sun2023aligning, zhou2024aligning, yu2023rlhf, xiao2024detecting}, we select LLaVA-1.5-7B~\citep{liu2023improved} and LLaVA-1.5-13B~\citep{liu2023improved} as the backbone models and apply \ours  on them.
The prompts used to generate preference data are randomly sampled from two categories, `complex\_reasoning\_77k' and `detail\_23k', in LLaVA's visual instruction tuning dataset, LLaVA-Instruct-150K, thus avoiding introducing additional data. 
We sample a total of 17k prompts.
To demonstrate the generalizability of \ours, we also choose VILA-7B~\cite{lin2024vila}, a recent LVLM, as the base model for our experiments. 
Similar to the LLAVA experimental setting, we randomly sample 17k prompts from the VILA visual instruction tuning dataset to generate preference pairs for training.
After obtaining the preference pairs, we finetune LLaVA and VILA with \ours on this data using LoRA~\citep{hu2021lora} for three epochs on LLaVA-1.5-7B, one epoch on LLaVA-1.5-13B, and one epoch on VILA-7B since we find that LLaVA-1.5-13B and VILA-7B is prone to overfitting on the sampled dataset.
All experiments are conducted on one A100 80GB GPU with 15 gpu hours for three epochs training on LLaVA-1.5-7B, 7 gpu hours for one epoch training on LLaVA-1.5-13B, and 6 gpu hours for one epoch training on VILA-7B.


\paragraph{Baselines} 
For the baselines, we compare with three previous methods that use preference optimization to improve LVLM performance: LLaVA-RLHF~\citep{sun2023aligning}, HA-DPO~\citep{zhao2023beyond}, and POVID~\citep{zhou2024aligning}. 
LLaVA-RLHF trains a reward model by incorporating additional human-annotated preference data and then finetunes LLaVA using PPO.
HA-DPO uses GPT to rewrite AI-generated responses for hallucination mitigation and data augmentation and then apply DPO to fine-tune the LVLM.
POVID introduces GPT to inject hallucinations into the ground truth answers and add noise to images to induce hallucinations in the LVLM to obtain negative samples and also uses DPO to finetune the LVLM.
These three methods are all based on LLaVA-1.5-7B.
Besides, we compare the method of using the ground truth answer as the positive sample and the LVLM-generated response as the negative sample for DPO finetuning, which we refer to as GT-DPO.
We also report comparison with other popular open-source LVLMs as a reference to demonstrate the superiority of our experimental results in Appendix~\ref{app: other lvlms}.

\paragraph{Benchmarks}
We select 14 hallucination and comprehensive benchmarks for evaluation. For the hallucination benchmark, we randomly sample 5000 images from the COCO~\citep{lin2014microsoft} validation set and randomly pair them with 5 questions, resulting in 5000 <image, question> pairs. We then evaluate the object hallucination rate on these 5000 pairs using the CHAIR~\citep{rohrbach2018object} metric
, calculated as follows:
\begin{equation}\label{eq: chair}
\resizebox{\hsize}{!}{$
\begin{aligned}
& \text{CHAIR}_I = \frac{|\{ \text{hallucinated objects} \}|}{|\{ \text{all mentioned objects} \}|}, \\
& \text{CHAIR}_S = \frac{|\{ \text{captions with hallucinated objects} \}|}{|\{ \text{all captions} \}|}.
\end{aligned}
$}
\end{equation}
We also use MM-Hal~\citep{sun2023aligning} and Mementos~\citep{wang2024mementos} as benchmarks for evaluating hallucination. In Mementos, we use F1 score as the metric to assess the LVLM's object hallucination and behavior hallucination when understanding multi-image inputs.
For the comprehensive benchmark, we select nine commonly used comprehensive benchmarks and general VQA tasks: LLaVA in the Wild~\citep{liu2024visual}, ScienceQA~\citep{lu2022learn}, TextVQA~\citep{singh2019towards}, MME Perception~\citep{fu2024mme}, MME Cognition~\citep{fu2024mme}, MMBench~\citep{liu2023mmbench}, MM-Vet~\citep{yu2023mm}, SeedBench~\citep{li2023seed}, and VisWiz~\citep{gurari2018vizwiz}. For details on these benchmarks, please refer to the Appendix~\ref{app: bench}.

\paragraph{Experiment results}
\textbf{(a) \ours can significantly reduce hallucinations of LVLMs.}
As shown in Table~\ref{exp: hallucination}, \ours significantly improves the performance of all three LVLMs on five hallucination benchmarks. On the CHAIRs, CHAIRi, and Mementos-Object benchmarks, which test object hallucination, \ours improves he performance of LLaVA-1.5-7B, LLaVA-1.5-13B, and VILA-7B by an average of 16.1\%, 7.1\%, and 8.4\%, respectively. 
On the MM-Hal benchmark, which uses GPT as an evaluator for a more comprehensive assessment of hallucinations, \ours achieves 12.7\%, 10.1\%, and 5.1\% performance improvement compared with LLaVA-1.5-7B, LLaVA-1.5-13B, and VILA-7B.
Notably, despite our three critic metrics focusing primarily on object hallucination, \ours also achieves the greatest improvement of 13.1\% on the Mementos-Behavior benchmark based on LLaVA-1.5-7B model, which tests behavior hallucination arising from understanding sequential image inputs. This improvement is significant because there is a correlation between object hallucination and behavior hallucination in sequential image understanding \citep{wang2024mementos}; reducing object hallucination increases the likelihood of correctly inferring the corresponding behavior.
\textbf{(b) \ours also enhances the comprehension capabilities of LVLMs.}
As shown in Table~\ref{exp: comprehensive}, on the nine comprehensive and VQA benchmarks, although the improvements are not as pronounced as on the hallucination benchmarks, \ours still achieves an average improvement of 3.5\%, 2.1\%, and 4.4\% compared to LLaVA-1.5-7B, LLaVA-1.5-13B, and VILA-7B. This is superior to other preference tuning methods.

\subsection{Importance of our critic metric}
\label{sec: Importance of our critic metric}

\begin{table*}[!htb]
\centering
\caption{The performance comparison between training LLaVA with preference pairs obtained using metric-inclusive and metric-free critic prompts in the in-context self-critic process.}
\label{exp: metrics}
\resizebox{\textwidth}{!}{ 
\begin{tabular}{l|ccccc|ccccccccc}
\toprule 
 & \multicolumn{5}{c}{Hallucination Benchmark} & \multicolumn{9}{c}{Comprehensive Benchmark} \\
\cmidrule(lr){2-6}\cmidrule(lr){7-15}
& CHAIRs & CHAIRi & MM-Hal & Mem$^{\mathrm{O}}$ & Mem$^{\mathrm{B}}$ & LLaVA$^{\mathrm{W}}$  & SQA$^{\mathrm{I}}$ & VQA$^{\mathrm{T}}$ & MME$^{\mathrm{P}}$ & MME$^{\mathrm{C}}$ & MMB & MM-Vet & SEED & VisWiz \\
\midrule
LLaVA-1.5-7B              & 50.8     & 11.7    & 2.04 & 39.29\% & 23.02\% & 63.4 & 66.8 & 58.2 & 1506.4 & 355.7 & 64.3 & 30.5 & 58.6 & 50.0\\
+ \ours w/o metrics        & 41.5     & 10.8    & 2.12 & 41.55\% & 23.92\% & 63.3 & 68.9 & 58.3 & 1504.6 & 371.7 & 64.0 & 31.5 & 60.4 & 53.7\\
+ \textbf{\ours} (ours)    & 40.9     & 10.4    & 2.30 & 46.08\% & 26.03\% & 66.1 & 69.1 & 58.5 & 1507.7 & 379.3 & 64.9 & 31.6 & 60.6 & 54.4\\
\bottomrule
\end{tabular}
}
\end{table*}

\begin{wrapfigure}[12]{R}{4cm}
\vspace{-10pt}
\includegraphics[width=4cm]{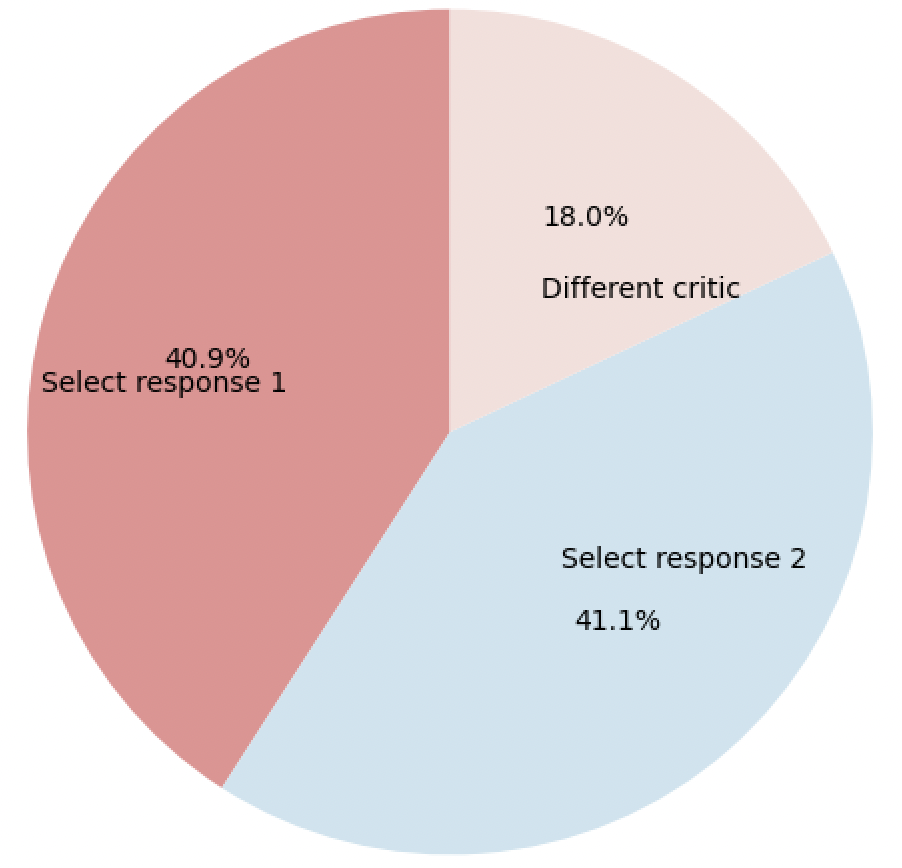}
\vspace{-15pt}
\caption{Comparison of critic results with and without critic metrics in \ours.}
\label{fig: metric_distribution}
\end{wrapfigure}

In this section, our main objective is to demonstrate the importance of the three critic metrics in the in-context self-critic stage through experiments and case studies. We use LLaVA-1.5-7B as base model to conduct experiments.
As in the experimental setup described in Section~\ref{sec: benchmark eval}, the prompts used to generate response candidates are sampled from LLaVA's visual instruction tuning dataset. After self-generating the response candidates, we keep these candidates unchanged and use LLaVA to evaluate them with both metric-inclusive and metric-free critic prompts, resulting in preference pairs that are then used to update the LLaVA.
We test the performance of both methods on 14 benchmarks, with the results shown in Table~\ref{exp: metrics}. Upon comparison, we find that removing the critic metrics still improved performance compared to the original LLaVA, but there remained a significant gap compared to \ours with metrics. 
This disparity is particularly notable in more challenging tasks like MM-Hal and Mementos, where the improvement from \ours without critic metrics is quite limited.
This demonstrates that with the correct design of critic prompts, LVLMs can gain critic capabilities and improve model performance without requiring instruction fine-tuning. Moreover, the three visual critic metrics are crucial for further enhancing performance.

\begin{table}[!htb]
\vspace{-5pt}
\centering
\caption{Comparison of response critic results with human judgment.}
\label{tab: human eval}
\vspace{-5pt}
\resizebox{\linewidth}{!}{%
\begin{tabular}{c|ccc}
\toprule 
& Select 1 & Select 2 & Align w. human  \\
\midrule
Human                 & 183  & 317  & -    \\
GPT-4v                & 198  & 302  & 95.6\% \\
\ours                 & 215  & 285  & 89.8\% \\
\ours w/o metrics     & 246  & 254  & 78.2\% \\
\bottomrule
\end{tabular}
}
\end{table} 

We compare the evaluation results distribution of response candidates with and without using metrics, as shown in Figure~\ref{fig: metric_distribution}. 
It can be seen that approximately 20\% of the response candidates have inconsistent evaluations between the two methods. 
Additionally, we randomly sample 500 response candidates and evaluate them both manually by the authors of this paper and with GPT-4v. 
For human evaluation, we provide 500 response pairs and asked individuals to directly select the better one. 
For GPT-4v, we use the same critic prompt with metrics as \ours for the evaluation.
Comparing these evaluations with SIMA's results in Table~\ref{tab: human eval}, we find that without the critic metrics, SIMA's evaluations are only 78\% consistent with human evaluations. 
After incorporating metrics, this consistency improved by 11.2\% to 89.8\%, which is very close to the evaluation results of GPT-4v and human.
In Appendix~\ref{app: prompt case study}, we also present an example of evaluation results with and without metrics to further illustrate the magic of these three visual metrics.



\subsection{Ablation studies}

\begin{figure}[!htbp]
\centering
\subfigure[{Average performance of LLaVA-1.5-13B with \ours at different iterations.}]{
\includegraphics[width=0.4\textwidth]{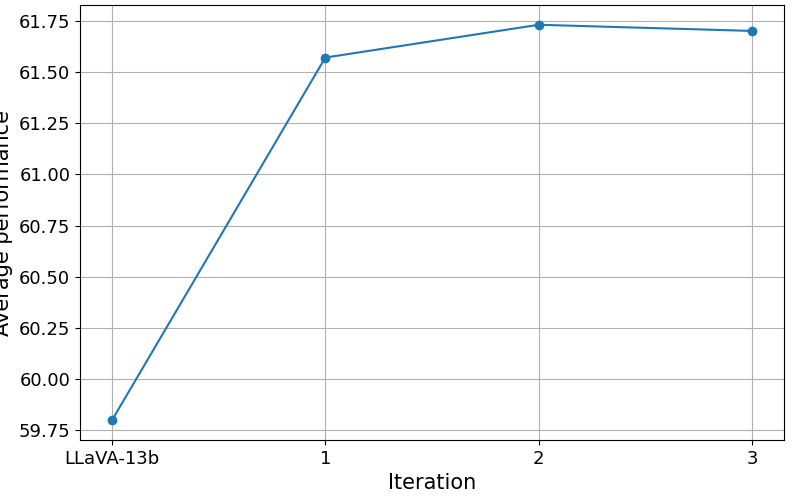}
\label{fig: ablation study_iter}
}
\hfil
\subfigure[{Average performance of \ours on LLaVA-1.5-7B with different decoding temperature.}]{
\includegraphics[width=0.4\textwidth]{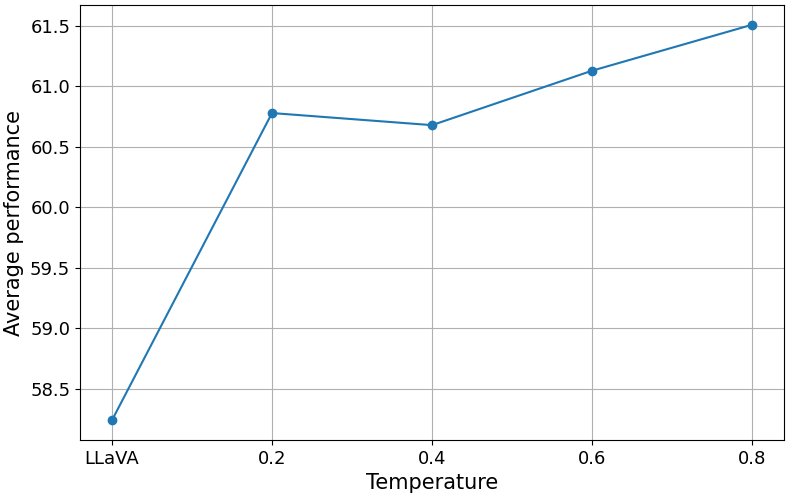}
\label{fig: ablation study_temp}
}
\vspace{-10pt}
\caption{{ Ablation studies of \ours.}}
\label{fig: ablation study}
\end{figure}

In this section, we conduct ablation studies on \ours from two aspects: the performance variation of \ours under multi-iteration finetuning and the impact of different decoding temperatures on performance when generating response candidates.

\paragraph{Performance of multi-iteration finetuning}
Figure~\ref{fig: ablation study_iter} shows the average performance of the model on benchmarks across different training iterations on LLaVA-1.5-13B. For detailed performance on each benchmark, please refer to Table~\ref{exp: iters_13B} in the Appendix~\ref{app: exp}. 
In each iteration, we randomly resample 17k prompts from LLaVA's visual instruction tuning dataset for self-generation. 
We observe that the performance improvement is most noticeable in the first iteration compared to the base model.
In the second iteration, there is an improvement, but it is not as pronounced. Although the average performance saturates in the third iteration, performance in some benchmarks continues to improve, as seen in Table~\ref{exp: iters_13B}.

\paragraph{Different decoding temperatures}
We also conduct an ablation study on the temperature used in temperature decoding during the response self-generation phase. The experimental results are shown in Figure~\ref{fig: ablation study_temp}. We find that as the temperature increases, the performance of SIMA also improves. We believe this is because, as the temperature increases, the responses generated by LVLM become more diverse and are more likely to exhibit hallucination. This increases the distribution shift between the responses generated by greedy decoding and those generated with higher temperature, leading to better performance improvements for LVLM during the preference tuning phase.

\section{Related Work}
\paragraph{Vision-Language Models}
Vision-Language Models (VLMs)~\cite{li2019visualbert, li2020oscar, wang2021simvlm, radford2021learning, li2022grounded} have emerged as critical tools in bridging visual and textual modalities, enabling advancements in multimodal understanding and reasoning tasks. 
Recent developments have been driven by the integration of large language models (LLMs)~\cite{touvron2023llama, jiang2023mistral, chiang2023vicuna} and sophisticated image encoders, leading to more robust and versatile Large Vision-Language Models (LVLMs)~\cite{bai2023qwen,zhu2023minigpt,chen2023sharegpt4v,dai2024instructblip,lin2024vila,yao2024minicpm,liu2024visual}. For instance, models like LLaVA~\cite{liu2024visual} and InstructBLIP~\cite{dai2024instructblip} combine advanced vision encoders with LLMs, enhancing their ability to follow vision-language instructions. 
In this work, we focus on further enhancing LVLM's visual understanding and reasoning abilities based on LVLM's visual instruction tuning data through self-improvement.

\paragraph{Modality Alignment}
Vision-language modality misalignment is a key challenge in LVLMs, where the generated textual outputs may not fully correspond to the input visual data.
Preference learning~\cite{rafailov2024direct, azar2024general, ethayarajh2024kto} is generaly used to improve modality alignment in LVLMs.
Some methods, such as using human annotation~\cite{sun2023aligning, yu2024rlhf} and third-party AI model feedback~\cite{li2023silkiepreferencedistillationlarge, zhao2023beyond, zhou2024aligning, yu2024rlaif, jing2024fgaif,xiong2024llava} for preference learning, have been proposed. 
However, these methods are resource-intensive and may introduce additional external hallucinations, leading to LVLM performance that is uncontrollable and unstable after optimization.
In this paper, we addresses both issues through a self-improvement approach, significantly enhancing modality alignment without introducing any external models or data.

\paragraph{Self-Improvement in Large Language Models}
Self-improvement is proposed in LLM to improve LLM itself with self-generated data. Several papers have explored self-improvement in LLM~\cite{yuan2024self, pang2024iterativereasoningpreferenceoptimization, wu2024metarewardinglanguagemodelsselfimproving,li-etal-2024-quantity, wang2024towards}.
To the best of our knowledge, this paper is the first to explore self-improvement in LVLMs.
Different from previous self-improvement methods in LLM which need to finetune the LLM with additional instruction tuning data before critic, our method demonstrate that LVLM can acquire the ability to act as a critic by properly configuring critic prompt without fine-tuning.

\section{Conclusion}
\label{sec: conclu}

In this paper, we introduce \ours framework in enhancing the alignment between visual and language modalities in LVLMs through self-improvement. 
This is achieved through self-generated responses, evaluating them via in-context self-critic, and employing preference tuning. 
\ours bypasses the need for the third-party AI model for data generation and response evaluation, making it more scalable and cost-effective.  This approach not only improves the modality alignment but also significantly enhances the model's comprehension abilities and reduces hallucinations across various benchmarks. 


\section*{Limitations}
One limitation of this paper is that the reliance on self-generated responses and self-critic inherently ties the \ours's performance to the current capabilities of LVLM and does not address the inherent potential biases caused by the vision instruction tuning dataset. 
This can result in \ours providing less significant improvements for LVLMs on certain benchmarks, such as LLaVA-1.5-7B and LLaVA-1.5-13B on TextVQA.
In future work, we will further explore this issue.

\section*{Broader Impacts}
\label{sec:impact}
To the best of our knowledge, we are the first to apply self-rewarding in LVLMs. This approach avoids the introduction of external models and data, enhancing the alignment between visual and language modalities through the model itself. This significantly reduces hallucinations and improves reasoning capabilities, greatly increasing the reliability of LVLMs.

From a societal impact perspective, while \ours has made substantial progress, it has not entirely eliminated potential risks within LVLMs. For example, reliance on self-generated and self-critic data may unintentionally reinforce biases caused by distribution shifts in the training data. Therefore, despite \ours's significant advancements, it is crucial to implement ethical guidelines and safeguards to mitigate these risks and ensure responsible use of this technology.

\section*{Acknowledgement}

Wang, Zhou and Huang are supported by DARPA Transfer from Imprecise and Abstract Models to Autonomous Technologies (TIAMAT) 80321, National Science Foundation NSF-IIS-2147276 FAI, DOD-ONR-Office of Naval Research under award number N00014-22-1-2335, DOD-AFOSR-Air Force Office of Scientific Research under award number FA9550-23-1-0048, DOD-DARPA-Defense Advanced Research Projects Agency Guaranteeing AI Robustness against Deception (GARD) HR00112020007, Adobe, Capital One and JP Morgan faculty fellowships.

\bibliography{neurips_2024}

\appendix
\section{Detailed critic prompt}
\label{App1: Detailed ranking prompt}
\begin{figure*}[!h]
\centering
\includegraphics[width=\textwidth]{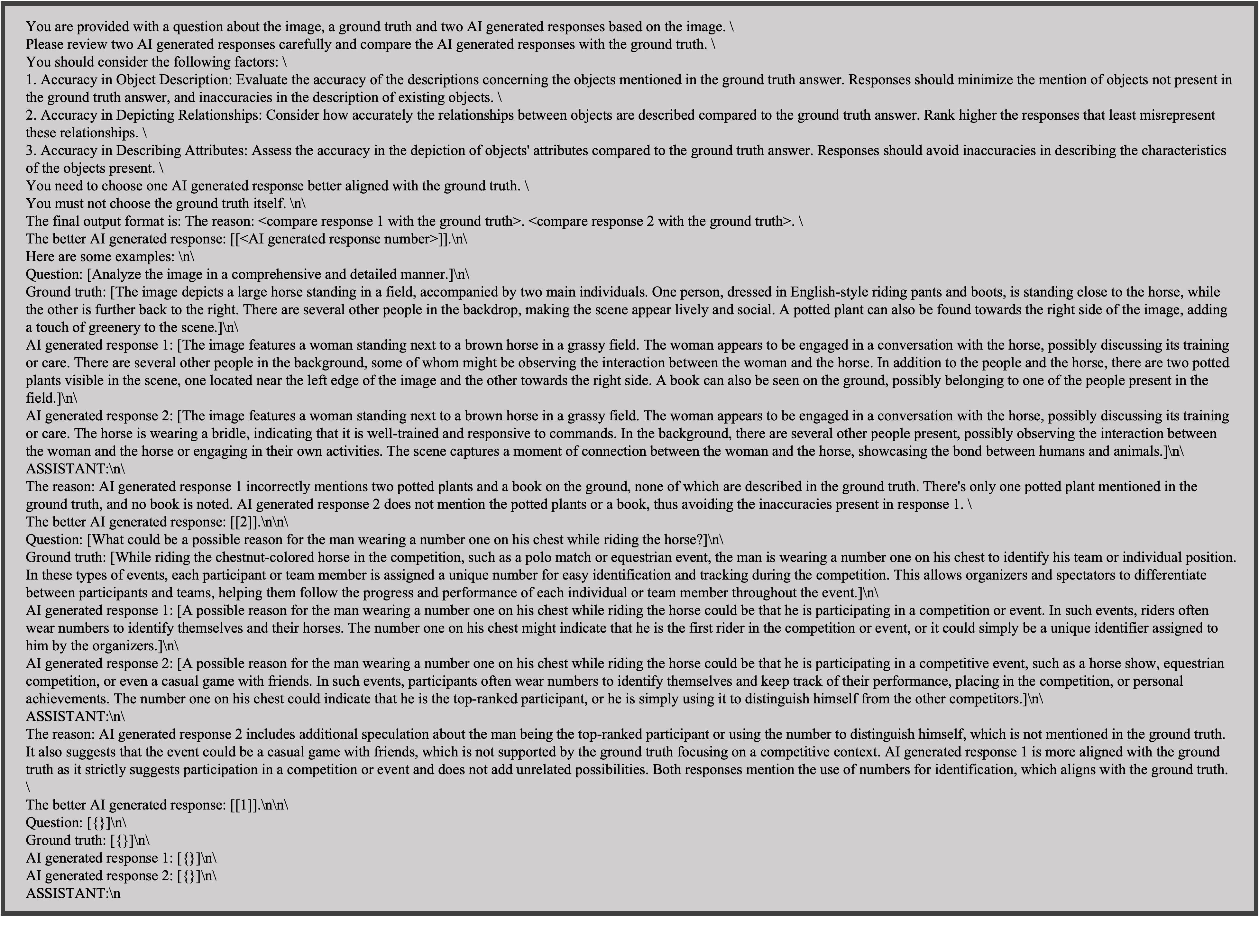}
\caption{Critic prompt used for in-context self-critic.}
\label{fig: ranking prompt}
\end{figure*}

In this section, we provide detailed critic prompt used during the in-context self-critic phase, as shown in Figure~\ref{fig: ranking prompt}.

\section{Detailed ablation studies}
\label{app: exp}

\subsection{LLaVA-1.5-7B}

In this section, we provide the performance of LLaVA-1.5-7B-7B across all benchmarks at different training epochs, as shown in Table~\ref{exp: epochs_7B}. Additionally, in Table~\ref{exp: temp}, we present the performance of \ours across all benchmarks when using different temperature coefficients for temperature decoding.

\begin{table*}[!htb]
\centering
\caption{Performance of different epochs on LLaVA-1.5-7B.}
\label{exp: epochs_7B}
\resizebox{\textwidth}{!}{ 
\begin{tabular}{l|ccccc|ccccccccc}
\toprule 
 & \multicolumn{5}{c}{Hallucination Benchmark} & \multicolumn{9}{c}{Comprehensive Benchmark} \\
\cmidrule(lr){2-6}\cmidrule(lr){7-15}
& CHAIRs & CHAIRi & MM-Hal & Mem$^{\mathrm{O}}$ & Mem$^{\mathrm{B}}$ & LLaVA$^{\mathrm{W}}$  & SQA$^{\mathrm{I}}$ & VQA$^{\mathrm{T}}$ & MME$^{\mathrm{P}}$ & MME$^{\mathrm{C}}$ & MMB & MM-Vet & SEED & VisWiz \\
\midrule
LLaVA-1.5-7B& 50.8     & 11.7   & 2.04 & 39.29\% & 23.02\% & 63.4 & 66.8 & 58.2 & 1506.4 & 355.7 & 64.3 & 30.5 & 58.6 & 50.0\\
+ \textbf{\ours} Epoch 1     & 43.9     & 10.8   & 2.17 & 42.39\% & 23.88\% & 65.3 & 68.9 & 58.2 & 1511.9 & 369.6 & 64.9 & 30.5 & 60.1 & 53.7\\
+ \textbf{\ours} Epoch 2     & 41.6     & 10.4   & 2.28 & 45.71\% & 24.93\% & 66.1 & 69.2 & 58.2 & 1514.8 & 371.8 & 65.0 & 31.5 & 60.4 & 54.0\\
+ \textbf{\ours} Epoch 3     & 40.9     & 10.4   & 2.30 & 46.08\% & 26.03\% & 66.1 & 69.1 & 58.5 & 1507.7 & 379.3 & 64.9 & 31.6 & 60.6 & 54.4\\
\bottomrule
\end{tabular}
}
\end{table*}

\begin{table*}[!htb]
\setlength{\abovecaptionskip}{0.2cm}
\centering
\caption{Performance of different decoding temperature.}
\label{exp: temp}
\resizebox{\textwidth}{!}{ 
\begin{tabular}{l|ccccc|ccccccccc}
\toprule 
 & \multicolumn{5}{c}{Hallucination Benchmark} & \multicolumn{9}{c}{Comprehensive Benchmark} \\
\cmidrule(lr){2-6}\cmidrule(lr){7-15}
& CHAIRs & CHAIRi & MM-Hal & Mem$^{\mathrm{O}}$ & Mem$^{\mathrm{B}}$ & LLaVA$^{\mathrm{W}}$  & SQA$^{\mathrm{I}}$ & VQA$^{\mathrm{T}}$ & MME$^{\mathrm{P}}$ & MME$^{\mathrm{C}}$ & MMB & MM-Vet & SEED & VisWiz \\
\midrule
T=0.2     & 40.2     & 10.1    & 2.11   & 45.42\% & 24.99\% & 65.2 & 68.5 & 58.3 & 1505.0 & 371.8 & 64.7 & 31.1 & 60.1 & 53.7\\
T=0.4     & 40.7     & 10.2    & 2.19   & 45.93\% & 25.37\% & 64.9 & 68.9 & 58.3 & 1506.4 & 355.7 & 65.0 & 31.1 & 60.3 & 53.8\\
T=0.6     & 40.9     & 10.3    & 2.23   & 45.71\% & 25.61\% & 65.7 & 69.2 & 58.2 & 1504.8 & 371.8 & 64.9 & 31.3 & 60.3 & 54.1\\
T=0.8     & 40.9     & 10.4    & 2.30   & 46.08\% & 26.03\% & 66.1 & 69.1 & 58.5 & 1507.7 & 379.3 & 64.9 & 31.6 & 60.6 & 54.4\\
\bottomrule
\end{tabular}
}
\end{table*}

\subsection{LLaVA-1.5-13B}
In this section, we present the detailed performance of LLaVA-1.5-13B across all benchmarks at different epochs and iterations in Tables~\ref{exp: epochs_13B} and Tables~\ref{exp: iters_13B}, respectively. It can be observed that the best results for LLaVA-1.5-13B are achieved after just one epoch. During multiple iteration training, performance on some benchmarks continues to improve in the third iteration while some declines due to overfitting.

\begin{table*}[!htb]
\centering
\caption{Performance of different epochs on LLaVA-1.5-13B.}
\label{exp: epochs_13B}
\resizebox{\textwidth}{!}{ 
\begin{tabular}{l|ccccc|ccccccccc}
\toprule 
 & \multicolumn{5}{c}{Hallucination Benchmark} & \multicolumn{9}{c}{Comprehensive Benchmark} \\
\cmidrule(lr){2-6}\cmidrule(lr){7-15}
& CHAIRs & CHAIRi & MM-Hal & Mem$^{\mathrm{O}}$ & Mem$^{\mathrm{B}}$ & LLaVA$^{\mathrm{W}}$  & SQA$^{\mathrm{I}}$ & VQA$^{\mathrm{T}}$ & MME$^{\mathrm{P}}$ & MME$^{\mathrm{C}}$ & MMB & MM-Vet & SEED & VisWiz \\
\midrule
LLaVA-1.5-13B & 48.6     & 10.8   & 2.19 & 40.37\% & 24.65\% & 66.5 & 71.6 & 61.3 & 1531.1 & 296.1 & 67.7 & 36.1 & 61.6 & 53.6\\
+ \textbf{\ours} Epoch 1       & 45.8     & 10.6   & 2.41 & 45.84\% & 27.17\% & 67.4 & 72.5 & 61.2 & 1538.1 & 298.6 & 68.4 & 38.3 & 63.0 & 55.5\\
+ \textbf{\ours} Epoch 2       & 46.1     & 10.6   & 2.26 & 45.53\% & 26.99\% & 67.2 & 72.4 & 61.2 & 1537.5 & 291.1 & 68.5 & 37.6 & 63.0 & 55.0\\
+ \textbf{\ours} Epoch 3       & 45.9     & 10.6   & 2.21 & 45.61\% & 26.74\% & 66.0 & 72.4 & 61.1 & 1529.2 & 291.4 & 68.3 & 35.9 & 63.0 & 54.9\\
\bottomrule
\end{tabular}
}
\end{table*}

\begin{table*}[!htb]
\centering
\caption{Performance of different iterations on LLaVA-1.5-13B.}
\label{exp: iters_13B}
\resizebox{\textwidth}{!}{ 
\begin{tabular}{l|ccccc|ccccccccc}
\toprule 
 & \multicolumn{5}{c}{Hallucination Benchmark} & \multicolumn{9}{c}{Comprehensive Benchmark} \\
\cmidrule(lr){2-6}\cmidrule(lr){7-15}
& CHAIRs & CHAIRi & MM-Hal & Mem$^{\mathrm{O}}$ & Mem$^{\mathrm{B}}$ & LLaVA$^{\mathrm{W}}$  & SQA$^{\mathrm{I}}$ & VQA$^{\mathrm{T}}$ & MME$^{\mathrm{P}}$ & MME$^{\mathrm{C}}$ & MMB & MM-Vet & SEED & VisWiz \\
\midrule
LLaVA-1.5-13B & 48.6     & 10.8   & 2.19 & 40.37\% & 24.65\% & 66.5 & 71.6 & 61.3 & 1531.1 & 296.1 & 67.7 & 36.1 & 61.6 & 53.6\\
+ \textbf{\ours} Iter 1        & 45.8     & 10.6   & 2.41 & 45.84\% & 27.17\% & 67.4 & 72.5 & 61.2 & 1538.1 & 298.6 & 68.4 & 38.3 & 63.0 & 55.5\\
+ \textbf{\ours} Iter 2        & 45.3     & 10.6   & 2.46 & 46.02\% & 27.58\% & 67.5 & 72.7 & 61.2 & 1528.9 & 298.6 & 68.5 & 38.3 & 62.9 & 55.9\\
+ \textbf{\ours} Iter 3        & 45.4     & 10.6   & 2.42 & 46.91\% & 27.63\% & 67.3 & 72.6 & 61.1 & 1529.8 & 298.6 & 68.6 & 37.9 & 63.0 & 56.0\\
\bottomrule
\end{tabular}
}
\end{table*}

\subsection{Comparison with other open-source LVLMs}
\label{app: other lvlms}
In this section, we report the performance of five other popular open-source LVLMs (BLIP-2~\citep{li2023blip}, InstructBLIP~\citep{dai2024instructblip}, IDEFICS~\citep{laurenccon2024obelics}, Qwen-VL-Chat~\citep{bai2023qwen}, and mPLUG-Owl2~\citep{ye2023mplug}) as a reference to demonstrate the superiority of our experimental results in Table~\ref{exp: other lvlms}. 
Compared to other open-source LVLMs, \ours also significantly outperforms all except for Qwen-VL-Chat on MM-Vet.

\begin{table*}[!htb]
\setlength{\abovecaptionskip}{0.2cm}
\centering
\caption{Performance comparison between \ours and other open-source LVLMs on comprehensive benchmarks}
\label{exp: other lvlms}
\resizebox{\textwidth}{!}{ 
\begin{tabular}{l|ccccccccc}
\toprule 

LVLMs & LLaVA$^{\mathrm{W}}$ $\uparrow$ & SQA$^{\mathrm{I}}$ $\uparrow$ & VQA$^{\mathrm{T}}$ $\uparrow$ & MME$^{\mathrm{P}}$ $\uparrow$ & MME$^{\mathrm{C}}$ $\uparrow$ & MMB $\uparrow$ & MM-Vet $\uparrow$ & SEED $\uparrow$ & VisWiz $\uparrow$ \\
\midrule
BLIP-2                  & 38.1 & 61.0  & 42.5  & 1293.8      & 290.0  & - & 22.4  & 46.4 &  19.6  \\
InstructBLIP            & 60.9 & 60.5  & 50.1  & 1212.8      & 291.8  & 36.0 & 26.2 & 53.4 & 34.5  \\
IDEFICS                 & 45.0 & -     & 25.9  & 1177.3      & -  & 30.0  & 30.0 & 45.0 & 35.5 \\
Qwen-VL-Chat            & 67.7 & 68.2  & 61.5  & 1487.6      & 360.7  & 60.6 & 47.3  & 58.2 & 38.9  \\
mPLUG-Owl2              & 59.9 & 68.7  & 58.2  & 1450.2      & 313.2  & 64.5 & 36.2  & 57.8 & 54.5  \\
\midrule
LLaVA-1.5-7B              & 63.4 & 66.8 & 58.2 & 1506.4 & 355.7 & 64.3 & 30.5 & 58.6 & 50.0\\
\rowcolor{gray!30}
+ \textbf{\ours} (ours)   & \textbf{66.1} & 69.1 & \textbf{58.5} & 1507.7 & \textbf{379.3} & \textbf{64.9} & \textbf{31.6} & \textbf{60.6} & \textbf{54.4}\\
\midrule
VILA-7B                   & 69.7	&68.2	&64.4	&1533.0	&316.4	&68.9	&34.9	&61.1	&57.8\\
\rowcolor{gray!30}
+ \textbf{\ours} (ours)   & \textbf{73.5}	&\textbf{72.2}	&\textbf{66.1}	&\textbf{1559.6}	&\textbf{326.8}	&\textbf{69.2}	&\textbf{38.4}	&\textbf{62.5}	&\textbf{62.1} \\
\bottomrule
\end{tabular}
}
\end{table*}

\section{Benchmark details}
\label{app: bench}

\textbf{LLaVA$^{\mathrm{W}}$} is an extensive benchmark for assessing visual reasoning models. It includes 24 varied images accompanied by a total of 60 questions, encompassing scenarios from indoor and outdoor settings to abstract art.

\textbf{ScienceQA}  is a multi-modal benchmark designed to evaluate and diagnose the multi-hop reasoning capabilities and interpretability of artificial intelligence systems in science. It provides an extensive data set of approximately 21,000 multiple-choice questions covering a wide range of scientific topics, supported by detailed answer notes, relevant lectures and explanations.

\textbf{TextVQA} is a dataset that benchmarks visual reasoning based on text in images. TextVQA requires models to read and reason about text in images to answer questions about them. Specifically, the model needs to incorporate a new form of text into the image and reason about it to answer the TextVQA question.

\textbf{MME} serves as a comprehensive benchmark for evaluating the capabilities of LVLMs in multimodal tasks. It evaluates models systematically across two main dimensions: perception and cognition, using 14 carefully designed subtasks that test the models' interpretative and analytical abilities.

\textbf{MMBench} introduces a two-pronged approach: a carefully curated dataset that significantly expands the scope and diversity of evaluation questions, and a groundbreaking CircularEval strategy that leverages ChatGPT to transform free-form predictions for structured choices.

\textbf{MM-Vet} is an evaluation benchmark specially designed to evaluate the multi-faceted capabilities of LVLM. It systematically builds complex multimodal tasks into 16 different ensembles derived from combinations of 6 core visual language features, providing granular analysis of model performance across different question types and answer styles.

\textbf{SEEDBench} is intended to rigorously assess the generative comprehension capabilities of LVLMs. It includes a comprehensive dataset of 19K multiple-choice questions with accurate human annotations, spanning 12 distinct evaluation dimensions that test both spatial and temporal understanding across image and video modalities.

\textbf{VizWiz} is a dataset in the field of visual question answering (VQA) derived from a naturalistic setting containing over 31,000 visual questions. It features a goal-oriented approach, featuring images taken by blind people, accompanied by their verbal queries, and crowdsourced answers.

\section{Hyperparameters}

In this section, we provide the hyperparameters used during training, as well as the GPT version utilized during evaluation, as shown in Table~\ref{tb: hyper}.

\begin{flushleft}
\begin{table}[!htbp]
\setlength{\abovecaptionskip}{0.2cm}
\centering
\caption{Hyperparameters of \ours during training and evaluation.}
\label{tb: hyper}
\begin{tabular}{l|l}
\toprule 
Parameter&Value \\
\midrule
lora r & 128  \\
lora alpha & 256  \\
mm projector lr & 2e-5  \\
learning rate & 1e-7  \\
model max length & 2048  \\
batch size & 1  \\
decoding temperature & 0.8 \\
GPT api version (Eval) & gpt-4-turbo \\
\bottomrule
\end{tabular}
\end{table}
\end{flushleft}
\section{Case Study}

\subsection{\ours case study}
\label{app: sima case study}
\begin{figure*}[!h]
\centering
\includegraphics[width=\textwidth]{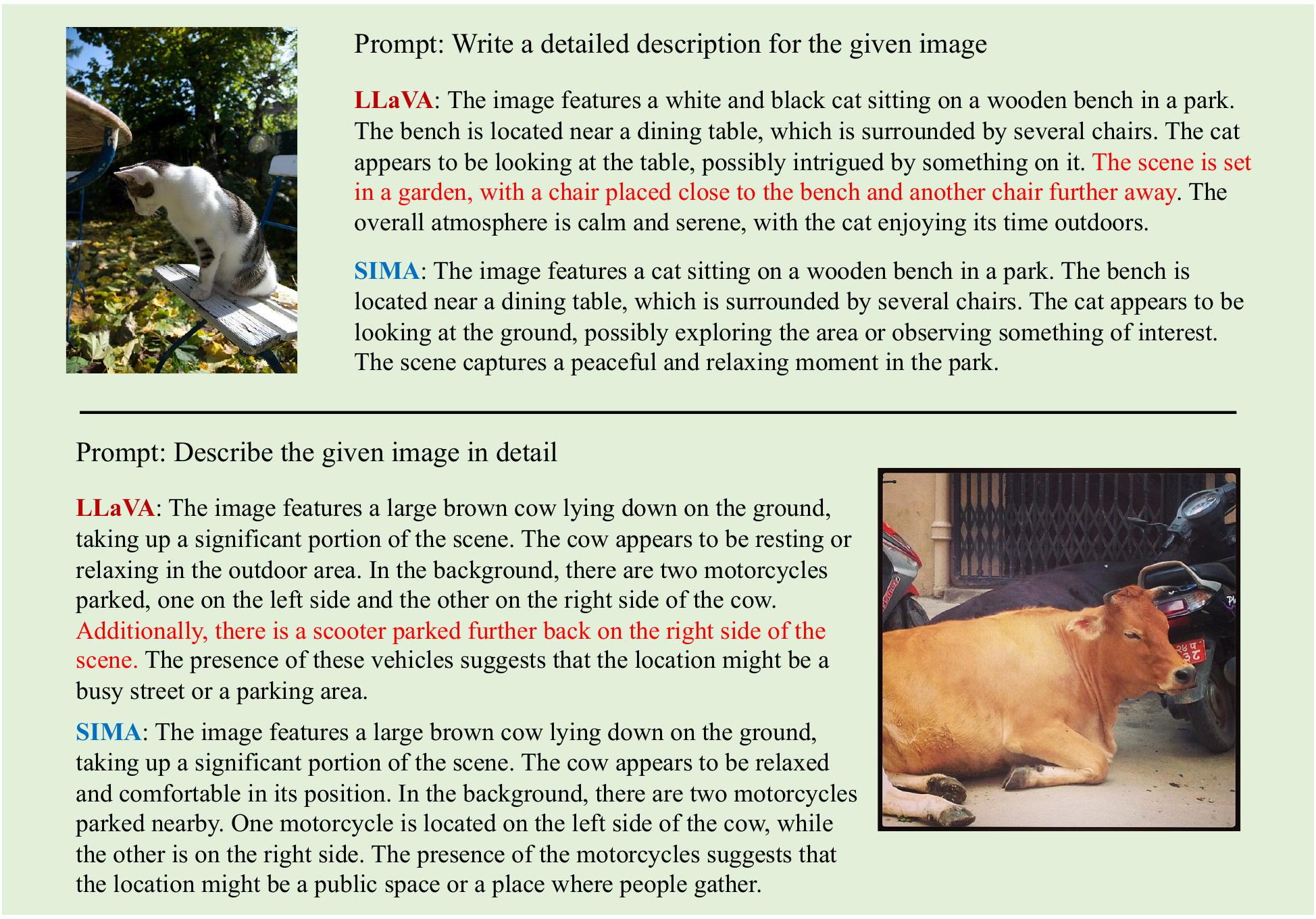}
\caption{Examples to illustrate the effect of \ours on LVLM. \ours is presented to be less hallucinated compared with LLaVA.}
\label{fig: demo_LLaVA}
\end{figure*}

In Figure \ref{fig: demo_LLaVA}, we compare the detailed captions generated from LLaVA and our method \ours. For instance, in the above demonstration in Figure \ref{fig: demo_LLaVA}, LLaVA inaccurately describes the scene `with a chair close to the bench and another chair further away', details that are not present in the provided image, while \ours accurately captures the details in images. Both demonstrations in Figure \ref{fig: demo_LLaVA} indicate that \ours is capable of generating more accurate descriptions with fewer hallucinations.

\subsection{Critic prompt case study}
\label{app: prompt case study}
\begin{figure*}[!h]
\centering
\includegraphics[width=0.95\textwidth]{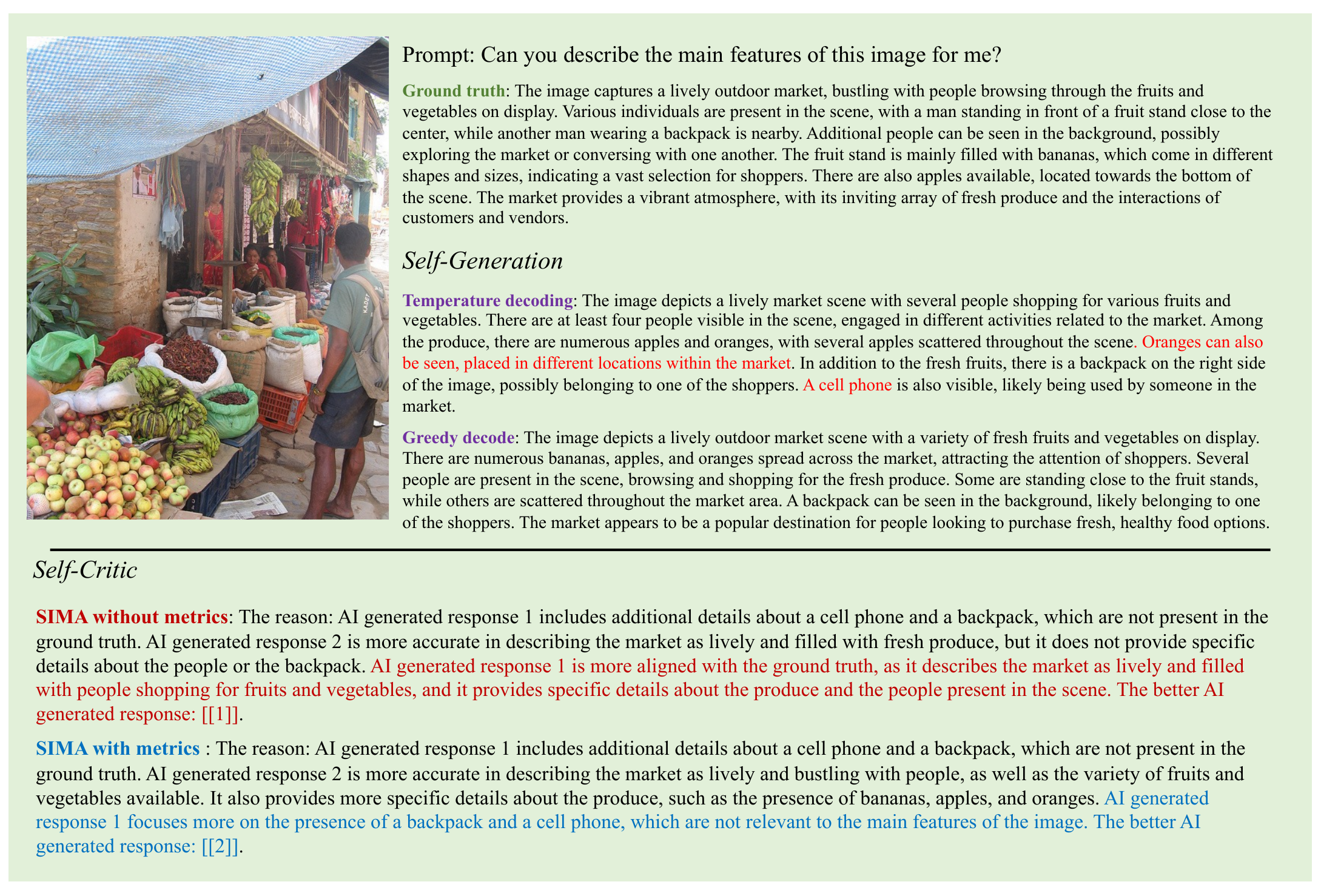}
\caption{An example of the critic results with and without using critic metrics in the in-context self-critic stage. The results show that after using metrics, the LVLM's evaluation is more focused on the details of the visual content, leading to a correct judgment.}
\label{fig: demo}
\end{figure*}

In Figure~\ref{fig: demo}, we present an example of evaluation results with and without metrics. The example shows that after using the critic metrics, the \ours's evaluation is guided to focus more on the details of the visual content, leading to correct judgments. Therefore, based on the analysis and results above, it is evident that critic metrics are crucial for improving the accuracy of response evaluations during in-context self-critic.

\section{Status of Exemption from Institutional Review Board}
\label{app: irb}
Before starting any segments of the study involving human evaluation, the research team completed and submitted a ``Human Subjects Research Determination" form to the appropriate Institutional Review Board (IRB). We obtained a determination letter from the IRB before any human study activities commenced, indicating that our project proposal had been granted `Exempt' status. This classification implies that the proposed research was deemed `Not Human Subjects Research'.

\end{document}